\documentclass{article}
\usepackage{spconf,amsmath,graphicx}

\usepackage{graphics} 
\usepackage{epsfig} 
\usepackage{mathptmx} 
\usepackage{times} 
\usepackage{amsmath} 
\usepackage{amssymb}  

\usepackage{amsfonts, dsfont}
\usepackage[inline]{enumitem}
\usepackage{pgfplots}
\usetikzlibrary{pgfplots.groupplots}
\usepackage{cite}
\usepackage{subcaption}
\usepackage{tikz}
\usetikzlibrary{calc}

\usepackage{array,booktabs,multirow}
\newcolumntype{M}[1]{>{\centering \arraybackslash}m{#1}}
\newcolumntype{L}[1]{>{\arraybackslash}m{#1}}
\newcommand {\tworow}[1]{\multirow{2}{*}{#1}}

\usepackage{hyperref}

\title{
Gumbel-NeRF: Representing Unseen Objects as\\ Part-Compositional Neural Radiance Fields
}

\name{Yusuke Sekikawa$^{\ddag\star}$ \qquad Chingwei Hsu$^{\dagger\star}$\thanks{${\star}$ Equal  contribution.}   \qquad Satoshi Ikehata$^{\dagger}$ \qquad Rei Kawakami$^{\dagger}$ \qquad Ikuro Sato$^{\dagger\ddag}$}
\address{$^{\dagger}$Tokyo Institute of Technology, Japan, ~~ $^{\star}$Denso IT Laboratory, Japan}

\begin{document}

\maketitle
\thispagestyle{empty}
\pagestyle{empty}

\begin{abstract}
We propose Gumbel-NeRF, a mixture-of-expert (MoE) neural radiance fields (NeRF) model with a hindsight expert selection mechanism for synthesizing novel views of unseen objects. 
Previous studies have shown that the MoE structure provides high-quality representations of a given large-scale scene consisting of many objects.
However, we observe that such a MoE NeRF model often produces low-quality representations in the vicinity of experts' boundaries when applied to the task of novel view synthesis of an unseen object from one/few-shot input.
We find that this deterioration is primarily caused by the foresight expert selection mechanism, which may leave an unnatural discontinuity in the object shape near the experts' boundaries.
Gumbel-NeRF adopts a hindsight expert selection mechanism, which guarantees continuity in the density field even near the experts' boundaries.
Experiments using the SRN cars dataset demonstrate the superiority of Gumbel-NeRF over the baselines in terms of various image quality metrics. \textit{The code will be available upon acceptance.}
\end{abstract}

\section{INTRODUCTION}
 Construction of 3D representations of unseen objects from 2D observations is important for various applications in robotics and autonomous driving, such as semantic mapping\cite{hosseinzadeh2019real,grinvald2019volumetric,sucar2020nodeslam}, obstacle avoidance\cite{tang2023rgb,ok2021hierarchical,lin2021multi} and scene understanding\cite{murthy2017reconstructing,lin2021multi}. One of the difficulties in this long-standing problem lies in capturing detailed properties of objects, including 3D-shape, texture, material, and reflectance. It becomes even more challenging due to ill-posedness, when only partial observations are available. As an illustrative example, methods of 3D representation construction for novel-view synthesis from one- or few-shot observations are still in high demand for an automotive application of 360-degree surrounding-view system that synthesizes a bird's-eye view near the ego vehicle. 

  Extensive studies have been conducted to construct 3D scene representations. Conventional geometric reconstruction methods\cite{schoenberger2016sfm, newcombe2011dtam} incrementally create explicit representations based on dense observations but struggle to recover unobserved regions. Approaches based on learned discrete representations can model high-level features\cite{yan2018learning,lombardi2019neural}, 
  but they tend to require high resolution and high computational cost. In contrast, works that use continuous representations implicitly reconstruct objects as learnable functions, providing the potential to capture fine details\cite{mescheder2019occupancy,saito2019pifu,sitzmann2019scene,mildenhall2021nerf}. 

   \begin{figure}[t]
      \centering
      \includegraphics[width=\linewidth,clip,]{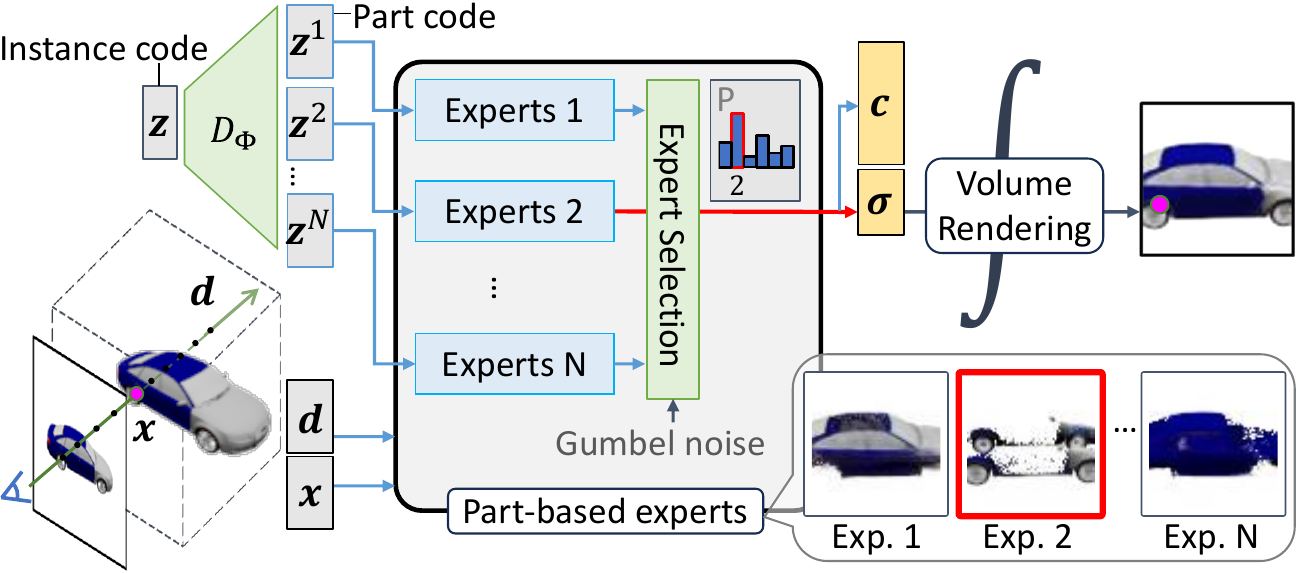}
      \caption{Overview of \textbf{Gumbel-NeRF}. 
      In the forward pass, a set of experts are processed to return densities and radiances.
      Out of $N$ experts, only one expert with the highest density is selected.
      This maximum-pooling expert selection guarantees continuity in the final density field, like the original NeRF.
      Each expert is associated with an expert-specific latent code so that the  expert learn to model a part of the object.}
      \label{fig:intro}
    \end{figure} 

Recently, Neural Radiance Field (NeRF)\cite{mildenhall2021nerf} has emerged as a 
milestone in the realm of continuous implicit representations, primarily designed for single, small-scale scenes. 
While NeRF is capable of generating remarkably high-quality synthesized images, it typically requires hundreds of images for training and must be optimized separately for each scene.
 
To overcome the constraints, CodeNeRF\cite{jang2021codenerf}, extended the original, scene-specific NeRF to model multiple and unseen instances of a semantic category. 
By conditioning on learnable instance-specific latent codes, 
CodeNeRF can synthesize novel views of an unseen instance even from highly limited input views. 
However, as CodeNeRF tries to encapsulate all of the properties into global latent codes using shared MLP, it sometimes struggles to adequately represent the variations of a semantic category when instances have diverse shapes and appearances. 

To enhance the expressivity of the NeRF models, Switch-NeRF\cite{zhenxing2022switch}, 
employed mixture-of-experts (MoE) structure to better represent large-scale scenes.
The MoE has a gate module that selects a single expert per input and the selected expert computes the density and radiance.
The set of experts, the gate module, and the other shared parameters are jointly optimized for a given scene and tested on the same scene.

In this work, we tackle the problem of high-quality novel-view syntheses for unseen car instances from one/few-shot input in the aim of enriching automotive applications.
A car is composed of \textit{parts}, such as wheels, roof, doors, \textit{etc.}, each of which is often visually similar to those in other cars.
Such a commonality would be well handled by a part-based formulation.
Intuitively, adopting a similar MoE structure to Switch-NeRE into CodeNeRF with part-specific latents appears to be a good solution for enhancing both expressivity and generalization.
However, we observed that the naive combination of the two results in a deterioration in the reconstructed shape around experts' boundaries (see the black car in Fig.~\ref{fig:qual-eval}-CSN).
We hypothesize that the primary cause of the deterioration comes from the \textit{foresight} expert selection mechanism; \textit{i.e.}, the gating network selects an expert \textit{before} processing the experts. 
Due to this gating mechanism, the best expert may not always be chosen in terms of rendering quality, thus it often results in unnatural discontinuity near the experts' boudaries.
We remark that discontinuity is an issue specific to novel view syntheses for unseen instances.

To address the above issue, we propose \textbf{Gumbel-NeRF}, a conditional MoE NeRF utilizing the \textit{hindsight expert selection mechanism} as depicted in Fig.~\ref{fig:intro}.
Gumbel-NeRF replaces the gate network in Switch-NeRF with the simple maximum pooling of the densities of the experts.
By this construction, Gumbel-NeRF guarantees continuity in the density field that the entire model returns, just like the original NeRF.
In addition, we introduce \textit{expert-specific codes} that represent different parts of a given car so that experts specifically learn to model the corresponding parts, whereas CodeNeRF uses a single code for a whole object.
We let the model learn how to decompose into parts, instead of giving supervision about parts.
Equipped with the enhanced expressivity and adaptability to test instances, 
Gumbel-NeRF outperforms the baselines in terms of several image quality metrics on a public benchmark of multi-instance view synthesis of cars.

    
\section{RELATED WORK}

   \begin{figure*}[thpb]
      \centering
      \begin{subfigure}{0.7\linewidth} 
            \includegraphics[width=\linewidth,clip,trim=20.5cm 7.5cm 20.5cm 7.5cm]{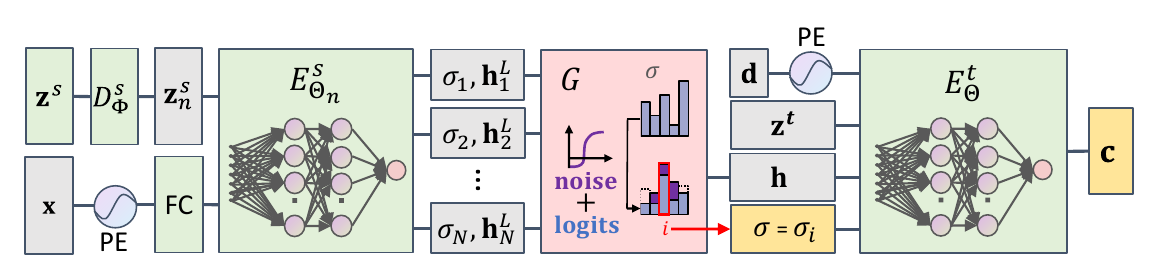}
            \caption{} \label{fig:arch}
        \end{subfigure}%
        \begin{subfigure}{0.3\linewidth}
            \centering
            \includegraphics[width=\linewidth,clip,trim=21cm 4.5cm 22cm 2.5cm]{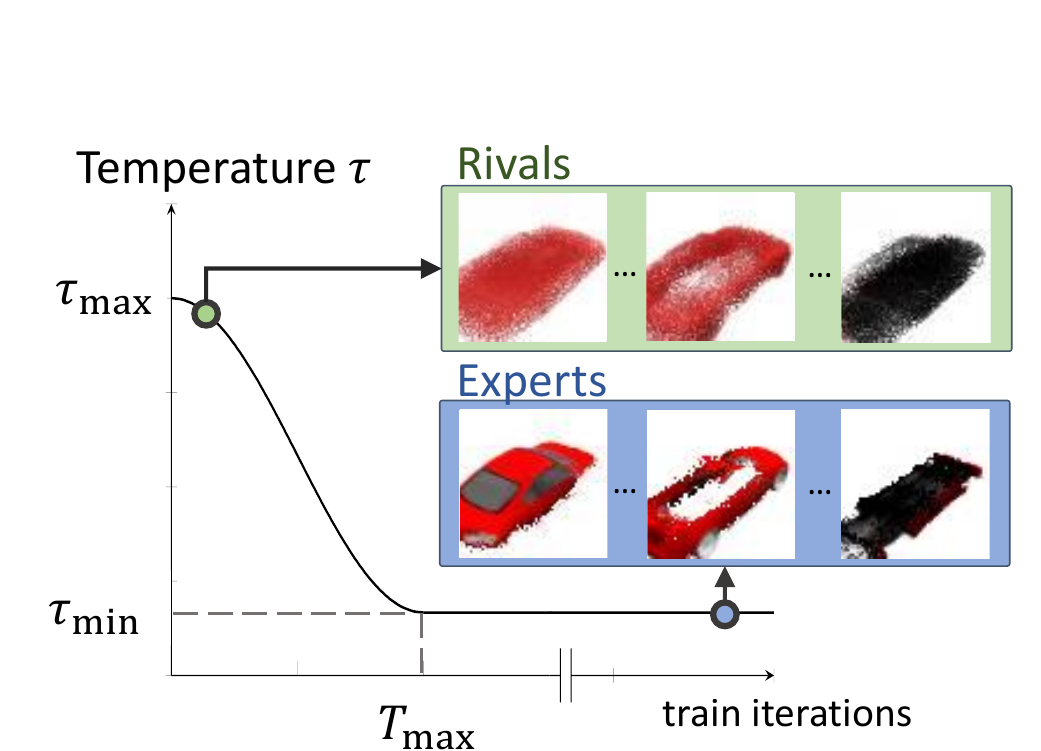} 
            \caption{} \label{fig:train}
        \end{subfigure}
        \caption{(a) Architecture of Gumbel-NeRF. Trainable parameters, output vectors, and our proposed expert selection layer are shown in green, yellow, and pink boxes, respectively. 
        FC refers to fully connected layer and PE refers to positional encoding. 
        $\textbf{z}^s_n$ denotes the expert-specific shape code for the $n$ -th expert.
        Different from Switch-NeRF\cite{zhenxing2022switch}, Gumbel-NeRF processes all experts in parallel to produce candidate densities $\sigma_{1\cdots N}$ and intermediate features $\textbf{h}^L_{1\cdots N}$.
        The layer $G$ samples the output from only one expert using the Gumbel-Max trick.
        (b) Scheduling of the temperature parameter.
        The temperature parameter used in the Gumbel-Max trick is scheduled to control the level of randomness through the training process.
        In the early stage, the temperature is set high so that all experts have a nearly equal chance of being selected (rival stage).
        In this stage, each expert obtains sufficient gradient updates, avoiding collapse (\textit{i.e.}, a vicious cycle where only one expert obtains all the gradient updates and other experts are underoptimized).
        Toward the end, the temperature is decreased to make experts distinct (expert stage).
        }
    \end{figure*}
\subsection{Neural Radiance Fields}
 NeRF\cite{mildenhall2021nerf} has achieved impressive success in the field of 3D scene representation and novel view synthesis. Using only multi-view supervision, NeRF models the volume density $\sigma$ and the emitted radiance $\textbf{c}$ of a scene by a continuous function $F_\Theta$ of 3D spatial coordinates $\textbf{x}$ and 2D viewing direction $\textbf{d}$ as
\begin{equation}
    F_\Theta: (\textbf{x, d}) \mapsto (\textbf{c}, \sigma). \label{eq:nerf}
\end{equation}
 This function is commonly approximated by MLPs. The constructed radiance fields can then be rendered into pixel values by volume rendering\cite{kajiya1984ray} given by
 \begin{equation} \label{eq:vr}
    \hat{C}(\textbf{r}) = \sum_{i=1}^{N_p}T_i (1-\exp{(-\sigma_i \delta_i)}) \textbf{c}_i,
\end{equation}
where $N_p$ is the number of sampled points along ray $\textbf{r}$, $\delta$ is the distance between two adjacent sample points, and
\begin{equation}
    T_n = \exp{\left(-\sum_{m=1}^{n-1}\sigma_m \delta_m\right)}
\end{equation}
can be interpreted as the accumulated transmittance up to sample $n$.
Comparing to discrete, explicit representations that are usually limited by resolution, the implicit nature of NeRF allows synthesizing photorealistic images from arbitrary viewpoints.
Despite the success of NeRF, one obvious drawback is that 
NeRF cannot handle test scenes with limited observations, since it was originally designed to model a single scene.

\subsection{Conditional NeRF} \label{section:cond-nerf}

A family of work\cite{jang2021codenerf,yu2021pixelnerf, reizenstein2021common,muller2022autorf,niemeyer2021giraffe,radford2021learning} extends the capabilities of NeRF to deal with multiple objects by conditioning the field representation with a set of tunable latent variables $\textbf{z}$ encoded with some prior knowledge. Adjusting $\textbf{z}$, a field representation can be controlled to represent different objects or scenes. A conditional NeRF can be generally formulated as:
\begin{equation}
F_\Theta(\textbf{x}, \textbf{d}, \textbf{z}) \mapsto (\textbf{c}, \sigma). \label{eq:cond-nerf}
\end{equation}
 The choice of encoding method and conditioning schemes varies across different works. CodeNeRF\cite{jang2021codenerf} jointly optimizes a set of instance-specific latent codes along with $\Theta$. During test time, given one or several images of a novel instance, CodeNeRF reconstructs its shape and synthesizes novel views by optimizing the instance-specific latent code. In contrast, PixelNeRF\cite{yu2021pixelnerf} directly obtains latents by extracting local features from input views with a CNN encoder, allowing conditioning on the rendering of the target view. Subsequent works\cite{reizenstein2021common} proposed different aggregation method of extracted local features. 
 AutoRF\cite{muller2022autorf} combines the characteristics of both approaches by employing an autoencoder to encode instance-specific latent codes. This enables image synthesis without test-time optimization, while also allowing refinement of image quality through such a process. Nevertheless, stronger supervision requirements including 3D bounding box and panoptic masks are needed.
 
\subsection{Decomposed NeRF} \label{section:decomp-nerf}
 Another family of work\cite{tancik2022block,zhenxing2022switch,tertikas2023partnerf} decomposes NeRF into multiple sub-models to enhance the network expressivity. 
 The optimization process of a NeRF model involves memorizing the entire scene, and as the scene size scales up, the synthesized image quality tends to deteriorate. Dividing the scene into distinct regions and assigning a neural network to each region can effectively address this problem.

 To achieve this decomposition, different approaches have been proposed. Block-NeRF\cite{tancik2022block} design hand-carft decomposition methods based on distance and street blocks, respectively. Another approach, Switch-NeRF\cite{zhenxing2022switch}, combines NeRF with a Sparsely-Gated Mixture of Experts (MoE)\cite{shazeer2017outrageously} framework, where the dispatchment of inputs to the experts (NeRFs) is determined by a jointly-optimized gating network. This approach improves the overall performance by reducing inconsistencies among sub-models 
 and better balancing the complexity of the scene.

 Decomposing NeRF also offers the benefit of enabling scene editing capabilities. GIRAFFE\cite{niemeyer2021giraffe} uses multiple generative NeRFs to model a scene with multiple objects, which allows for explicit control over individual objects in a scene, offering a disentangled manner of manipulation. 
 PartNeRF\cite{tertikas2023partnerf}, decomposes an object category into several semantic parts. Each part is handled by a separate NeRF network, which is trained in its own local coordinates. They employ complicated loss terms to ensure reasonable division of objects. We note that shape editing is not our goal.

\section{METHOD}

 As shown in Fig.~\ref{fig:intro}, our proposed method follows the basic formulation of conditional NeRFs given in Eq.~\ref{eq:cond-nerf}, which targets in modeling multiple instances within the same semantic category. 
 Following CodeNeRF\cite{jang2021codenerf}, our method jointly optimizes the neural radiance field $F_\Theta$, the latent code mapper $D_\Phi$, and a set of instance-specific latent codes $\{\textbf{z}_{m}=(\textbf{z}^s_{m}, \textbf{z}^t_{m})\}_{m=1}^M$, where $M$ denotes the number of instances in the training set and $(\textbf{z}^s_m, \textbf{z}^t_m)$ represent the shape and texture latent codes, respectively. 
 At test time, we optimize the instance-specific latent codes with frozen $F_\Theta$ and $D_\Phi$.

Fig.~\ref{fig:arch} illustrates the architecture of our proposed approach, which consists of two key elements:
(i) a set of $N$ experts, each of which couples to corresponding expert-specific latent codes, and
(ii) an expert selection mechanism that selects a single expert for a given input.
To effectively train experts, we introduce
(iii) a rival-to-expert training strategy that controls the level of randomness through the training so that different experts grow to model different parts of objects (see Fig.~\ref{fig:train}).
These elements are described in detail below.
   
\subsection{Part-Specific Experts}

 Similarly to Switch-NeRF\cite{zhenxing2022switch} and PartNeRF\cite{tertikas2023partnerf}, the neural implicit representation $F_\Theta$ consists of a set of $N$ submodels $\{E_{\Theta_n}\}_{n=1}^N$, which we refer to as experts. In addition to the original 5D inputs (location $\textbf{x}$ and viewing direction $\textbf{d}$), each expert $E_{\Theta_n}$ is exclusively conditioned on part-specific latent codes $\textbf{z}_{m,n} = (\textbf{z}^s_{m,n}, \textbf{z}^t_{m,n})$, associated to the $m$-th instance and to $n$-th expert. Here, $N$ is a predefined constant.
 
 The experts' design is adopted from that of the vanilla NeRF\cite{mildenhall2021nerf}, such that $E_{\Theta_n}$ is composed of two parts: a shape MLP ${E^s_{\Theta_n}}$ and a texture head ${E^t_{\Theta_n}}$. 
 The density $\sigma$ depends only on $\textbf{x}$, while the radiance $\textbf{c}$ depends on both $\textbf{x}$ and $\textbf{d}$. 
 Following Switch-NeRF, the texture code and the texture head for predicting the final outputs are designed to be shared across experts; namely, $\textbf{z}^t_{m} = \textbf{z}^t_{m,n}, {E^t_\Theta} = {E^t_{\Theta_n}}, n=1,\cdots,N$. 
 
 To obtain the expert-specific latent codes, we utilize a latent code mapper $D_\Phi$ that includes separate mappings for shape $D_\Phi^s$ and texture $D_\Phi^t$. The shape mapping consists of $N$ linear functions that map the instance-specific shape codes $\textbf{z}^s_{m}$ to $N$ expert-specific shape codes $\{\textbf{z}^s_{m,n}\}_{n=1}^N$ associated to corresponding experts. 
 As for the texture codes, since a unified RGB head is used, we let the texture mapping $D_\Phi^t$ be simply an identity function. 
 Formally, the latent code mappers are given as:
\begin{align}
     D_\Phi^s&: \textbf{z}^s_{m} \mapsto \{\textbf{z}^s_{m,n}\}_{n=1}^N, \\
     D_\Phi^t&:  \textbf{z}^t_{m} \mapsto \textbf{z}^t_{m}, \qquad m=1,\dots,M.
\end{align}

The expert parameters and the corresponding code are optimized in a co-adaptive manner so that the expert-code pair well represents a specific part of the target object. 
We will empirically show that experts can learn to decompose objects into similar parts without explicit supervision of parts. 
We omit the subscript $m$ for simplicity in the rest of the paper.

\subsection{Density-Based Expert Selection}
\label{section:exp-select}

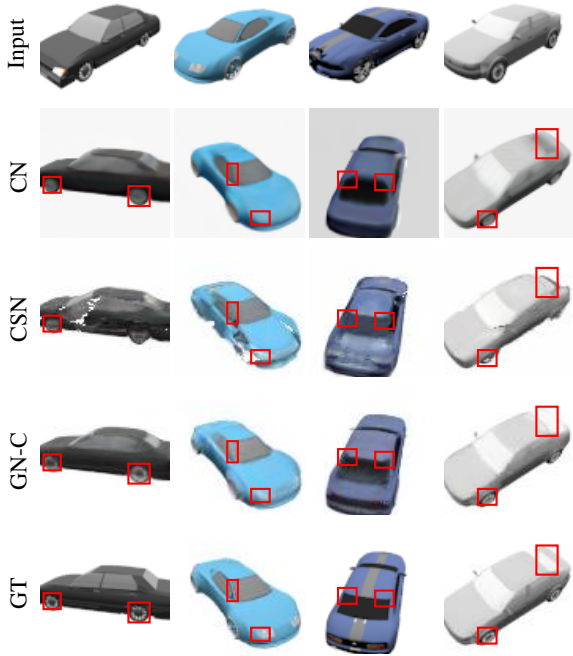
\begin{figure}[b!]
    \setlength{\tabcolsep}{1pt}
    \begin{tabular}{@{} M{0.05\linewidth} M{0.20\linewidth} M{0.20\linewidth} M{0.20\linewidth} M{0.20\linewidth} @{}} 
        \rotatebox[origin=c]{90}{\small Input}
        & \includegraphics[width=1\linewidth,clip,trim=8mm 8mm 8mm 8mm]{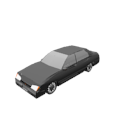}
        & \includegraphics[width=1\linewidth,clip,trim=8mm 8mm 8mm 8mm]{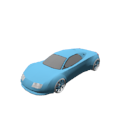}
        & \includegraphics[width=1\linewidth,clip,trim=8mm 8mm 8mm 8mm]{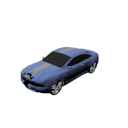}
        & \includegraphics[width=1\linewidth,clip,trim=8mm 8mm 8mm 8mm]{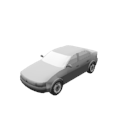}\\
        \rotatebox[origin=c]{90}{\small CN}
        & \begin{subfigure}[b]{1\linewidth}
            \centering
            \begin{tikzpicture}
              \node[anchor=south west,inner sep=0] (image) at (0,0) {\includegraphics[width=1\linewidth,clip,trim=8mm 8mm 8mm 8mm]{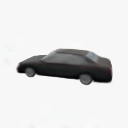}};
              \draw[red,thick] (1.5px,23px) rectangle (8px,17px);
              \draw[red,thick] (33.5px,19.5px) rectangle (41.5px,12px);
            \end{tikzpicture}
        \end{subfigure} 
        & \begin{subfigure}[b]{1\linewidth}
            \centering
            \begin{tikzpicture}
              \node[anchor=south west,inner sep=0] (image) at (0,0) {\includegraphics[width=1\linewidth,clip,trim=8mm 8mm 8mm 8mm]{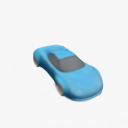}};
              \draw[red,thick] (20px,20px) rectangle (24px,28px);
              \draw[red,thick] (29px,5px) rectangle (36px,10px);
            \end{tikzpicture}
        \end{subfigure} 
        & \begin{subfigure}[b]{1\linewidth}
            \centering
            \begin{tikzpicture}
              \node[anchor=south west,inner sep=0] (image) at (0,0) {\includegraphics[width=1\linewidth,clip,trim=8mm 8mm 8mm 8mm]{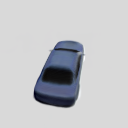}};
              \draw[red,thick] (11px,19px) rectangle (18px,25px);
              \draw[red,thick] (25px,18px) rectangle (32px,24px);
            \end{tikzpicture}
        \end{subfigure} 
        & \begin{subfigure}[b]{1\linewidth}
            \centering
            \begin{tikzpicture}
              \node[anchor=south west,inner sep=0] (image) at (0,0) {\includegraphics[width=1\linewidth,clip,trim=8mm 8mm 8mm 8mm]{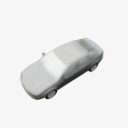}};
              \draw[red,thick] (13px,4px) rectangle (20px,10px);
              \draw[red,thick] (35px,30px) rectangle (43px,41px);
            \end{tikzpicture}
        \end{subfigure} \\
        \rotatebox[origin=c]{90}{\small CSN}
        & \begin{subfigure}[b]{1\linewidth}
            \centering
            \begin{tikzpicture}
              \node[anchor=south west,inner sep=0] (image) at (0,0) {\includegraphics[width=1\linewidth,clip,trim=8mm 8mm 8mm 8mm]{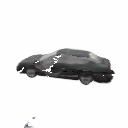}};
              \draw[red,thick] (1.5px,23px) rectangle (8px,17px);
            \end{tikzpicture}
        \end{subfigure} 
        & \begin{subfigure}[b]{1\linewidth}
            \centering
            \begin{tikzpicture}
              \node[anchor=south west,inner sep=0] (image) at (0,0) {\includegraphics[width=1\linewidth,clip,trim=8mm 8mm 8mm 8mm]{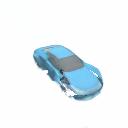}};
              \draw[red,thick] (20px,20px) rectangle (24px,28px);
              \draw[red,thick] (29px,5px) rectangle (36px,10px);
            \end{tikzpicture}
        \end{subfigure} 
        & \begin{subfigure}[b]{1\linewidth}
            \centering
            \begin{tikzpicture}
              \node[anchor=south west,inner sep=0] (image) at (0,0) {\includegraphics[width=1\linewidth,clip,trim=8mm 8mm 8mm 8mm]{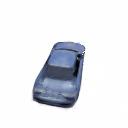}};
              \draw[red,thick] (11px,19px) rectangle (18px,25px);
              \draw[red,thick] (25px,18px) rectangle (32px,24px);
            \end{tikzpicture}
        \end{subfigure} 
        & \begin{subfigure}[b]{1\linewidth}
            \centering
            \begin{tikzpicture}
              \node[anchor=south west,inner sep=0] (image) at (0,0) {\includegraphics[width=1\linewidth,clip,trim=8mm 8mm 8mm 8mm]{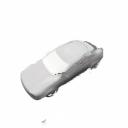}};
              \draw[red,thick] (13px,4px) rectangle (20px,10px);
              \draw[red,thick] (35px,30px) rectangle (43px,41px);
            \end{tikzpicture}
        \end{subfigure}\\
        \rotatebox[origin=c]{90}{\small GN-C}
        & \begin{subfigure}[b]{1\linewidth}
            \centering
            \begin{tikzpicture}
              \node[anchor=south west,inner sep=0] (image) at (0,0) {\includegraphics[width=1\linewidth,clip,trim=8mm 8mm 8mm 8mm]{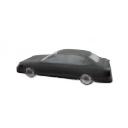}};
              \draw[red,thick] (1.5px,23px) rectangle (8px,17px);
              \draw[red,thick] (33.5px,19.5px) rectangle (41.5px,12px);
            \end{tikzpicture}
        \end{subfigure} 
        & \begin{subfigure}[b]{1\linewidth}
            \centering
            \begin{tikzpicture}
              \node[anchor=south west,inner sep=0] (image) at (0,0) {\includegraphics[width=1\linewidth,clip,trim=8mm 8mm 8mm 8mm]{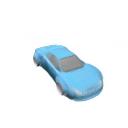}};
              \draw[red,thick] (20px,20px) rectangle (24px,28px);
              \draw[red,thick] (29px,5px) rectangle (36px,10px);
            \end{tikzpicture}
        \end{subfigure} 
        & \begin{subfigure}[b]{1\linewidth}
            \centering
            \begin{tikzpicture}
              \node[anchor=south west,inner sep=0] (image) at (0,0) {\includegraphics[width=1\linewidth,clip,trim=8mm 8mm 8mm 8mm]{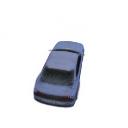}};
              \draw[red,thick] (11px,19px) rectangle (18px,25px);
              \draw[red,thick] (25px,18px) rectangle (32px,24px);
            \end{tikzpicture}
        \end{subfigure} 
        & \begin{subfigure}[b]{1\linewidth}
            \centering
            \begin{tikzpicture}
              \node[anchor=south west,inner sep=0] (image) at (0,0) {\includegraphics[width=1\linewidth,clip,trim=8mm 8mm 8mm 8mm]{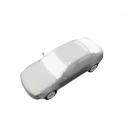}};
              \draw[red,thick] (13px,4px) rectangle (20px,10px);
              \draw[red,thick] (35px,30px) rectangle (43px,41px);
            \end{tikzpicture}
        \end{subfigure}\\
        \rotatebox[origin=c]{90}{\small GT}
        & \begin{subfigure}[b]{1\linewidth}
            \centering
            \begin{tikzpicture}
              \node[anchor=south west,inner sep=0] (image) at (0,0) {\includegraphics[width=1\linewidth,clip,trim=8mm 8mm 8mm 8mm]{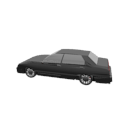}};
              \draw[red,thick] (1.5px,23px) rectangle (8px,17px);
              \draw[red,thick] (33.5px,19.5px) rectangle (41.5px,12px);
            \end{tikzpicture}
        \end{subfigure} 
        & \begin{subfigure}[b]{1\linewidth}
            \centering
            \begin{tikzpicture}
              \node[anchor=south west,inner sep=0] (image) at (0,0) {\includegraphics[width=1\linewidth,clip,trim=8mm 8mm 8mm 8mm]{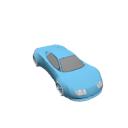}};
              \draw[red,thick] (20px,20px) rectangle (24px,28px);
              \draw[red,thick] (29px,5px) rectangle (36px,10px);
            \end{tikzpicture}
        \end{subfigure} 
        & \begin{subfigure}[b]{1\linewidth}
            \centering
            \begin{tikzpicture}
              \node[anchor=south west,inner sep=0] (image) at (0,0) {\includegraphics[width=1\linewidth,clip,trim=8mm 8mm 8mm 8mm]{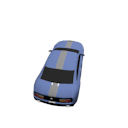}};
              \draw[red,thick] (11px,19px) rectangle (18px,25px);
              \draw[red,thick] (25px,18px) rectangle (32px,24px);
            \end{tikzpicture}
        \end{subfigure} 
        & \begin{subfigure}[b]{1\linewidth}
            \centering
            \begin{tikzpicture}
              \node[anchor=south west,inner sep=0] (image) at (0,0) {\includegraphics[width=1\linewidth,clip,trim=8mm 8mm 8mm 8mm]{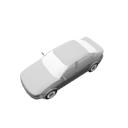}};
              \draw[red,thick] (13px,4px) rectangle (20px,10px);
              \draw[red,thick] (35px,30px) rectangle (43px,41px);
            \end{tikzpicture}
        \end{subfigure}\\
    \end{tabular}
    \caption[Qualitative results of novel view synthesis of unseen objects]{Qualitative results of novel view synthesis of unseen objects using one-shot test-time optimization. Compared to CodeNeRF (CN) and Coded Switch-NeRF (CSN), our Gumbel-NeRF (GN-C) generally produces higher quality, especially for those parts marked by red boxes.} \label{fig:qual-eval}
\end{figure}

Gumbel-NeRF adopts the hindsight-based density-based expert selection rule, as opposed to the foresight gating network adopted in Switch-NeRF\cite{zhenxing2022switch}. 
The use of a gating network to select an expert before processing experts might seem reasonable for our task setting; however, we have observed that the foresight gating mechanism often degrades the performance of the model, as is also evident in other literature\cite{royer2023revisiting}. 
The MoE design in Switch-NeRF prevents the gating network from sharing expert information, resulting in a solely location-based expert selection mechanism. 
The design is well-suited for a single, large-scale scene. 
This is because the complexity of a single scene is fixed so that Switch-NeRF can efficiently decompose the scene into regions, each of which can then be treated by vanilla NeRF reconstructions. 
However, in our task setting, we argue that the capacity of experts may be better leveraged when they specialize in handling regions with similar shape and appearance properties, not locations, across many instances.
Moreover, the MoE design in the Switch-NeRF does not guarantee continuity in the predicted density field, as opposed to the original NeRF.
Especially for the task of novel view syntheses of unseen instances, we observed an unnatural discontinuity in the predicted density field, resulting in severe deterioration of synthesizing quality (see the black car in Fig.~\ref{fig:qual-eval}-CSN).
 
To overcome the above-mentioned issues, we devised to place the expert selection \textit{after} the expert blocks. 
With this construction, the expert selection relies on as much information provided by the experts as possible. 
In our proposed method, the given inputs pass through all the shape MLPs, and only one of the outputs contributes to the final neural field. This process is formulated as:
\begin{align}
         E_{\Theta_n}^s &\;:\;  (\textbf{x},\,\textbf{z}^s_n) \mapsto (\textbf{h}^L_n, \sigma_n), \; n=1,\dots,N  \\
        G  &\;:\; \{(\sigma_n,\,\textbf{h}^L_n)\}_{n=1}^N \mapsto (\sigma, \, \textbf{h}),
    \end{align}
where $\textbf{h}$ are intermediate features. 
The expert selector $G$ basically adopts the maximum pooling of the densities returned by the experts; that is, an expert having the highest density is only used and the other experts are ignored.
With this construction, the model guarantees continuity in the predicted density field.
To address router collapse, in which a particular expert is always chosen, we employ the Gumbel-Max trick\cite{gumbel1948statistical}.
Namely, Gumbel noises are added to the output densities to control the chance rate for the $n$-th expert to be selected. 
We first treat $\{\sigma_n\}_{n=1}^N$ as the unnormalized ``probabilities" and calculate the normalized log probabilities using a LogSoftmax function:
    \begin{equation}
        \textbf{logits}_n = \log\left(\frac{\exp{(\frac{\log \sigma_n}{\tau})}}{\sum_{i=1}^{N} \exp{(\frac{\log \sigma_i}{\tau})}}\right), \; n=1,\dots,N,  \label{eq:softmax}
    \end{equation}
where $\tau$ is the temperature parameter to control the level of randomness, which will be further discussed in Sec.~\ref{section:train}.
 Then, the output is selected by:
    \begin{align}
        \sigma &= \mathds{1}_{\max}(\textbf{logits} + \textbf{g}) \cdot (\sigma_1, \sigma_2,\dots,\sigma_N), \\
        \textbf{h}^L &= \mathds{1}_{\max}(\textbf{logits} + \textbf{g}) \cdot (\textbf{h}^L_1, \textbf{h}^L_2,\dots,\textbf{h}^L_N), \label{eq:combine-h}
    \end{align}
where 
    \begin{equation}
    \begin{split}
         \mathds{1}_{\max}(\textbf{y})&=\delta_{n, \underset{i}{\text{argmax}}(\textbf{y}_i)}^\top  \\
        &= 
        \begin{cases}
          1 & \text{if } n=i\\
          0 & \text{otherwise}
        \end{cases}, \; n=1,\dots,N
    \end{split}
    \end{equation}
is a one-hot vector having 1 in the index corresponding to the maximum value of a vector $\textbf{y}$, and $\textbf{g}$ is a vector of $N$ i.i.d. Gumbel noise samples drawn from the standard Gumbel distribution using the inverse transform sampling technique:
    \begin{equation}
        g _j \sim -\log{(-\log{(\text{Uniform}(0, 1))})},\; j=1,\dots,N.
    \end{equation}

\subsection{Rival-to-Expert Training Strategy} \label{section:train}

 The temperature parameter $\tau>0$ in Eq.~\ref{eq:softmax} plays a crucial role in determining the level of randomness of $G$. When $\tau$ increases, the Softmax approaches the uniform distribution, resulting in nearly random sampling.
 In contrast, a smaller $\tau$ encourages a more consistent distribution, where logits maintain their original ranking and are less affected by the added Gumbel noise. 
 
 During training, we schedule $\tau$ using a cosine annealing followed by a constant final value as
    \begin{equation} \label{eq:temp_schedule}
        \tau(t) = \begin{cases}
            \tau_{\min} + \frac{\tau_{\max} - \tau_{\min}}{2} \left(1 + \cos {\frac{\pi \, t}{T_{\max}}}\right) \, &\text{if } t \leq T_{\max} \\
            \tau_{\min} \, &\text{otherwise},
        \end{cases}
    \end{equation}
where $\tau_{\max}, \tau_{\min}, T_{\max}$ are the initial temperature, final temperature, and the duration of cosine annealing in terms of the percentage of the total training iterations, respectively (See Fig.~\ref{fig:train}).
 By this scheduling, during the earlier stage, a higher level of randomness is introduced in selecting the experts. We refer to this stage as the \textit{rival stage}, where all experts have a similar chance of being selected and acquiring gradient updates. In this stage, the experts act as rivals, collectively modeling the overall scene in a coarse manner. As the temperature approaches $\tau_{\min}$, the training progresses to the \textit{ expert stage}. During this stage, the experts become ``real experts," as the selection process reaches a more stable state. This stability enables them to focus exclusively on the regions of interest and continuously refine the quality of reconstruction throughout the rest of the training process. This rival-to-expert training scheme effectively prevents the router collapse problem and training instabilities without the use of additional loss terms. The entire model is trained end-to-end by minimizing the photometric loss:
    \begin{equation}
        \underset{\Theta, \Phi, \{\textbf{z}_{m}\}_{m=1}^M}{\text{min}} \; \sum_{\textbf{r}\in \mathcal{R}} \lVert \hat{C}(\textbf{r}) - C(\textbf{r}) \rVert^2,
    \end{equation}
where $\hat{C}$, $C$ and $\mathcal{R}$ are the rendered pixel value, the ground truth pixel value and the set of rays in a training batch, respectively.

\section{EVALUATION}

\subsection{Datasets}

We conduct experiments on a re-rendered version of the ``car" category of ShapeNet\cite{chang2015shapenet} dataset provided by SRN\cite{sitzmann2019scene}. ShapeNet contains 3,514 instances of cars (2,458 for training, 704 for testing, and 352 for validation). Each training instance accompanies 50 images rendered from random viewpoints, while each testing instance accompanies 251 images rendered from the same set of predefined camera poses on an Archimedean spiral. All images contain a single foreground car against a white background. 

\subsection{Baselines}

 We compare our method, Gumbel-NeRF, with CodeNeRF\cite{jang2021codenerf}, which is also capable of handling multi-instance datasets with only 2D supervision. To show the effectiveness of our gate-free, density-based expert selection design, we also compare it with ``\textit{Coded Switch-NeRF}", a naive extension of Switch-NeRF\cite{zhenxing2022switch} where their gating network and experts are conditioned on latent codes. The components of Coded Switch-NeRF can be written as:
    \begin{align}
         G& \;:\; \left(\textbf{x}, \textbf{z}^s;\; \{\textbf{z}^s_n, F_{\Theta_n}\}_{n=1}^N\right) \mapsto \left( \textbf{z}^s_{n^\ast}, F^s_{\Theta_{n^\ast}} \right) \\ 
         {F^s_{\Theta_{n^\ast}}}& \;:\;  (\textbf{x},\, \textbf{z}^s_{n^\ast}) \mapsto \textbf{h}\; .
    \end{align}

\subsection{Evaluation Metrics}

 We report the results with standard metrics for evaluating image quality: PSNR, SSIM\cite{wang2004image} and the VGG, AlexNet and SqueezeNet versions of LPIPS\cite{zhang2018unreasonable}.

\begin{figure*}[]
    \centering
    \setlength{\tabcolsep}{1pt}
    \begin{tabular}{@{} M{0.05\linewidth} M{0.10\linewidth} M{0.10\linewidth} M{0.10\linewidth} M{0.10\linewidth} M{0.05\linewidth} M{0.10\linewidth} 
    M{0.10\linewidth}
    M{0.10\linewidth} M{0.10\linewidth} @{}} 
        & \small Expert 1 & \small Expert 2 & \small Expert 3 & \small Expert 4 & &\small Expert 1 & \small Expert 2 & \small Expert 3 & \small Expert 4 \\
        \rotatebox[origin=c]{90}{\small CSN}
        & \includegraphics[width=1\linewidth,clip,trim=6mm 6mm 6mm 6mm]{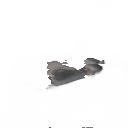}
        & \includegraphics[width=1\linewidth,clip,trim=6mm 6mm 6mm 6mm]{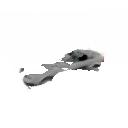}
        & \includegraphics[width=1\linewidth,clip,trim=6mm 6mm 6mm 6mm]{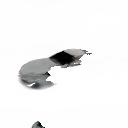}
        & \includegraphics[width=1\linewidth,clip,trim=6mm 6mm 6mm 6mm]{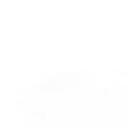}
        & &\includegraphics[width=1\linewidth,clip,trim=6mm 6mm 6mm 6mm]{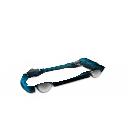}
        & \includegraphics[width=1\linewidth,clip,trim=6mm 6mm 6mm 6mm]{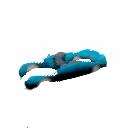}
        & \includegraphics[width=1\linewidth,clip,trim=6mm 6mm 6mm 6mm]{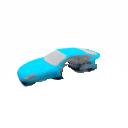}
        & \includegraphics[width=1\linewidth,clip,trim=6mm 6mm 6mm 6mm]{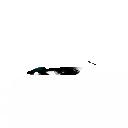}\\
        \rotatebox[origin=c]{90}{\small GN}
        & \includegraphics[width=1\linewidth,clip,trim=6mm 6mm 6mm 6mm]{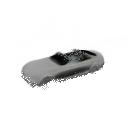}
        & \includegraphics[width=1\linewidth,clip,trim=6mm 6mm 6mm 6mm]{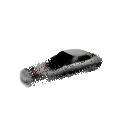}
        & \includegraphics[width=1\linewidth,clip,trim=6mm 6mm 6mm 6mm]{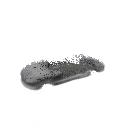}
        & \includegraphics[width=1\linewidth,clip,trim=6mm 6mm 6mm 6mm]{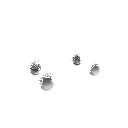}
        & &\includegraphics[width=1\linewidth,clip,trim=6mm 6mm 6mm 6mm]{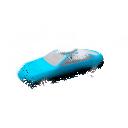}
        & \includegraphics[width=1\linewidth,clip,trim=6mm 6mm 6mm 6mm]{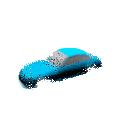}
        & \includegraphics[width=1\linewidth,clip,trim=6mm 6mm 6mm 6mm]{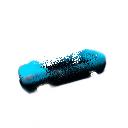}
        & \includegraphics[width=1\linewidth,clip,trim=6mm 6mm 6mm 6mm]{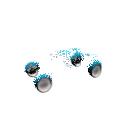}\\
    \end{tabular}
    \caption{Visualization of the decomposition provided by Coded Switch-NeRF (CSN) and Gumbel-NeRF (GN). Images in each column are rendered from only the 3D points handled by the corresponding expert.} \label{fig:decomp}
\end{figure*}

\subsection{Implementation Details}

 We implement Gumbel-NeRF and Coded Switch-NeRF with $N=4$ experts based on the released code of Switch-NeRF\cite{zhenxing2022switch}. We modify CodeNeRF's MLPs as the Gumbel-NeRF's expert architecture when comparing with CodeNeRF, but use half of its channels per layer to keep the number of parameters the same. Compared with Coded Switch-NeRF, the architecture of experts follows that of Switch-NeRF. We use 256-dimensional instance-specific codes and 128/256-dimensional part-specific codes for CodeNeRF-style and Switch-NeRF-style experts, respectively. 
 
 We follow the training and test-time optimization setups of CodeNeRF\cite{jang2021codenerf}, with carefully selected hyper-parameters to provide a fair comparison. Specifically, we train the models on 128 NVIDIA P100 GPUs with a total batch size of 327,680 points for 25k iterations using centered-cropped images and another 20k iterations using uncropped images. For test-time optimization, we iteratively refine only the instance-specific latent codes for 200 iterations. 
 
 During training, the learning rate starts with an initial value of $1.3 \times 10^{-3}$ and exponentially decays with a decay factor of 0.1 for cropped images, then remains constant for uncropped images. During test-time optimization, the same exponential decay scheduler is used, but with an initial learning rate of $2.0 \times 10^{-2}$. The models are trained with AdamW optimizer. 
 The parameters of Gumbel-NeRF cosine annealing are set as $\tau_{\max} =10, \tau_{\min} = 0.5$, and $T_{\max} = 20\%$ for training and as constant $\tau_{\min}$ for testing time optimization. \\

\begin{table}[]
\caption{Quantitative evaluation on ShapeNet-SRN cars test set. Note that we clip the rendered values to 0-1.}
\small
\label{tab:eval-test}
\begin{center}
    \begin{tabular}{L{8mm}M{8mm}M{8mm}M{8mm}M{8mm}M{8mm}M{8mm}}     
        \toprule
        \tworow{Method} & 
        \tworow{\shortstack{ \# of \\ params}} &
        \tworow{PSNR$\uparrow$} & 
        \tworow{\centering SSIM$\uparrow$} & 
        \multicolumn{3}{c}{LPIPS$\downarrow$} \\ 
            \cmidrule{5-7} & & & & VGG & Alex & Squeeze \\
        \midrule
        CN     & 0.7M & 19.66 & 0.882 & 0.150 & 0.161 & 0.119 \\
        GN-C   & 0.8M & \textbf{21.51} & \textbf{0.892} & \textbf{0.119} & \textbf{0.138} & \textbf{0.104} \\
        \midrule
        CSN    & 4.1M & 19.50 & 0.864 & 0.145 & 0.160 & 0.110 \\
        GN-S   & 3.9M & \textbf{21.43} & \textbf{0.890} & \textbf{0.114} & \textbf{0.126} & \textbf{0.088} \\
    \bottomrule
    \end{tabular}
\end{center}
\end{table}

\section{RESULTS}

\noindent\textbf{Reconstruction of Unseen Objects.}
 We perform an evaluation on the ShapeNet-SRN car test set, following the SRN\cite{sitzmann2019scene} and CodeNeRF\cite{jang2021codenerf}, which calculates the average values of each metric using images except those used as test-time optimization input. We conducted a one-shot optimization using only one input image from the same viewpoints across instances. 
 We show the quantitative and qualitative results of CodeNeRF (\textbf{CN})\cite{jang2021codenerf}, Coded Switch-NeRF (\textbf{CSN}), and Gumbel-NeRF (\textbf{GS-C} and \textbf{GS-S}, where ``-C'' and ``-S'' refers to CodeNeRF-style and Switch-NeRF-style expert architecture, respectively) in Table~\ref{tab:eval-test} and Fig.~\ref{fig:qual-eval}. As can be seen, with a similar number of parameters, Gumbel-NeRF outperforms Code-NeRF in all metrics.
 Our proposed method also outperforms Coded Switch-NeRF, showing the effectiveness of our gate-free density-based expert selection design despite the trade-off in processing time. 
 We can visually inspect that our model has no artifacts in the experts' boundary, while Coded Switch-NeRF has an unnatural discontinuity in shape.
 
\noindent\textbf{Part Decomposition.}
 We visualize the decomposition of objects in Fig.~\ref{fig:decomp} using images rendered from the 3D points handled by each expert. As can be seen, the Gumbel-NeRF is capable of learning a more consistent decomposition across objects compared to Coded Switch-NeRF. 
 For example, expert 4 of our Gumbel-NeRF generates 4 wheels for the left and right instances, while expert 4 of Coded Switch-NeRF generates almost nothing for the left instance and generates some lower components for the right instance.
 The inconsistent decomposition by Coded Switch-NeRF is due to the expert selection mechanism, in which the expert selection is done before ``seeing" the expert's quality.
 Furthermore, Gumbel-NeRF demonstrates a better ability to prevent the router collapse problem, as it evenly utilizes all experts in the training phase.





\section{CONCLUSION}
We proposed Gumbel-NeRF, a conditional neural radiance field capable of constructing the 3D representations of unseen car instances from one/few 2D observations. 
Our proposed method leverages the hindsight expert selection mechanism, which guarantees the continuous transition in the predicted density field, successfully increasing the expressibility of latent code-based NeRF. 
We also propose a novel rival-to-expert training strategy in order to balance the utilization of experts.
Through experiments on the ShapeNet-SRN cars dataset, we demonstrate that our method outperforms CodeNeRF and Coded Switch-NeRF in terms of several image quality metrics, proving its superior adaptability in capturing the details of unseen instances. 

\noindent\textbf{Acknowledgement.}
This work is supported by JSPS KAKENHI JP22H03642 and DENSO IT LAB Recognition and Learning Algorithm Collaborative Research Chair.


\small
\bibliographystyle{IEEEbib}
\bibliography{reference}

\end{document}